\documentclass[10pt,twocolumn,letterpaper]{article}

\usepackage{titlesec}
\usepackage{iccv}
\usepackage{times}
\usepackage{epsfig}
\usepackage{graphicx}
\usepackage{amsmath}
\usepackage{amssymb}
\usepackage{mathtools}
\usepackage{microtype}
\usepackage{import}
\usepackage{wrapfig}
\usepackage{xcolor}
\usepackage{layouts}
\usepackage{mathtools}
\usepackage{printlen}
\usepackage{balance}
\usepackage{sepfootnotes}
\usepackage[accsupp]{axessibility}
\graphicspath{{./graphics/}}
\usepackage[pagebackref=true,breaklinks=true,letterpaper=true,colorlinks,bookmarks=false]{hyperref}

\def\iccvPaperID{xxxx}

\ificcvfinal\fi

\newif\ifsupp
\suppfalse

\iccvfinaltrue

\newcommand{\surf}{\Gamma}

\newcommand{\energy}{I}

\newcommand{\image}{\mathcal{I}}

\newcommand{\pixel}{\mathbf{u}}

\newcommand{\R}{\mathbb{R}}

\newcommand{\eqn}[1]{\begin{align}#1\end{align}}

\newcommand{\spl}[1]{\begin{split}#1\end{split}}

\newcommand{\x}{\mathbf{x}}

\newcommand{\y}{\mathbf{y}}

\newcommand{\n}{n}

\newcommand{\back}{g_B}

\newcommand{\fore}{g_F}

\newcommand{\visibility}{\chi}

\newcommand{\overbar}[1]{\mkern 1.5mu\overline{\mkern-1.5mu#1\mkern-1.5mu}\mkern 1.5mu}

\definecolor{r}{HTML}{EF625C}
\definecolor{y}{HTML}{E3B67D}

\newtheorem{result}{Result}

\newcommand{\res}[1]{\begin{result}#1\end{result}}

\begin{document}
\title{A Theory of Topological Derivatives for Inverse Rendering of Geometry}

\author{Ishit Mehta \qquad Manmohan Chandraker \qquad Ravi Ramamoorthi \\
\\
University of California San Diego}

\maketitle

\begin{abstract}
    We introduce a theoretical framework for differentiable surface evolution
    that allows discrete topology changes through the use of topological
    derivatives for variational optimization of image functionals. While prior
    methods for inverse rendering of geometry rely on silhouette gradients for
    topology changes, such signals are sparse. In contrast, our theory derives
    topological derivatives that relate the introduction of vanishing holes and
    phases to changes in image intensity. As a result, we enable differentiable
    shape perturbations in the form of hole or phase nucleation. We
    validate the proposed theory with optimization of closed curves in 2D and
    surfaces in 3D to lend insights into limitations of current methods and
    enable improved applications such as image vectorization, vector-graphics
    generation from text prompts, single-image reconstruction of shape ambigrams
    and multiview 3D reconstruction.

\end{abstract}

\section{Introduction}
\begin{figure}
\begin{minipage}[c]{\textwidth}
    \footnotesize
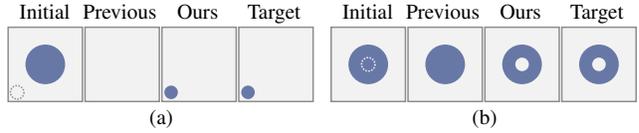
\end{minipage}
    \caption{\textbf{Pathologies of shape derivatives.} Previous work on inverse
    surface reconstruction relies on shape derivatives which are inadequate for
    instances when (a) the target geometry is far from the initialization, and when (b)
    target has a hole in the interior. We resolve these failure cases with
    topological derivatives.}
    \label{fig:teaser}
\end{figure}

Recovering geometry from images is a central theme for several problems in
vision and graphics, where a common approach is to derive the differential of
the rendering functional.
Depending on the type of surface representation used, such as B\'{e}zier
paths~\cite{liDifferentiableVectorGraphics2020},
triangle meshes~\cite{delaunoyGradientFlowsOptimizing2011,li18} or
level-sets~\cite{gargalloMinimizingReprojectionError2007}, corresponding gradient flow equations are
derived.
These works formulate image differentials as shape derivatives, which we posit
to be restrictive for inverse problems, since deformations induced by shape
derivatives (SD) are limited to surface boundaries.
This may lead to local minima when recovering geometry with high-genus topology.
In Figure~\ref{fig:teaser} we illustrate two inverse problems where shape
derivatives do not suffice to recover the optimal shape.
In such cases the optimization is required to make updates far away from the
boundary by (a) nucleating additional volume in the exterior, and (b)
perforating the interior of the shape.
Our work theoretically characterizes such shape perturbations in the form of topological derivatives (TD) for inverse rendering.

We derive TDs following the definition proposed by Sokolowski
and Zochowski~\cite{sokolowski1999topological}.
In the case of planar curves, vanishing balls are introduced in their interior or exterior.
In the limit, the TD is estimated as the difference in the image functionals for the perturbed
and unperturbed shapes.
The resulting gradient updates can prompt hole and phase nucleation in regions of high-error (see Figure~\ref{fig:teaser}). 
For closed surfaces in 3D, we introduce TDs with respect to conic perturbations through the interior of the visible shape.
We observe that visibility terms in SDs are evaluated only on the apparent contours of the visible shape.
In contrast, gradient flows using TDs can encourage visibility changes in the interior, resulting in a more robust and accurate recovery.

Informed by recent successes in inverse
rendering~\cite{bangaru2022differentiable,mehta2022level,vicini2022differentiable,wang2021neus,yariv2021volume},
we use level-sets~\cite{osher88} for surface representation.
We build on the extensive literature on \emph{variational}
level-sets~\cite{faugerasVariationalPrinciplesSurface,jin2003variational,solemGeometricFormulationGradient2005,zhao1996variational}
for surface reconstruction.
Our approach differs from the more recent methods on differentiable
rasterization~\cite{liDifferentiableVectorGraphics2020} and
rendering~\cite{bangaru2022differentiable,mitsuba3,li18,vicini2022differentiable}
that use differential calculus to derive geometry gradients.
The variational framework lends us a common structure to analyze the motion of continuous surfaces in both 2D and 3D.
It also provides a natural way to extend shape derivatives to the formulation for topological derivatives.
Additionally, using this approach, we are able to draw theoretical insights across literature spanning 3D reconstruction~\cite{carpioTopologicalDerivativesShape2008,gargalloMinimizingReprojectionError2007}, topology optimization~\cite{sokolowski1999topological}, differentiable vector graphics~\cite{liDifferentiableVectorGraphics2020} and differentiable rendering~\cite{mitsuba3,laine20,li18,ravi20}.

We structure this work as three theoretical results interspersed with
empirical observations for closed curves in 2D (\S~\ref{sec:2d}) and surfaces in
3D (\S~\ref{sec:3d}).
Previous work on differentiable rasterization~\cite{liDifferentiableVectorGraphics2020} proposes geometry gradients for standard vector graphics representations.
In Result~\ref{res:res1} (\S~\ref{sec:2d_sd}), we derive a corresponding shape derivative for level-sets in the variational setting.
Motivated by the above limitations, we derive topological derivatives for hole and phase nucleation in Result~\ref{res:res2} (\S~\ref{sec:3d_td}).
We show practical applications of image vectorization and text-to-vector graphics using an evolution equation that works with arbitrary loss functions.
In \S~\ref{sec:3d_sd} we reason about the necessity of TDs in 3D.
Result~\ref{res:res3} (\S~\ref{sec:3d_td}) discusses the construction of conic perturbations and the derivation for TD.
We show improvements over previous methods that rely on SDs in terms of speed for visibility optimization (Figure~\ref{fig:vis_shade_gradients}), accuracy of recovery for complex topology (Figure~\ref{fig:inverse_main}) and also show an application of reconstructing shape ambigrams from a single image (Figure~\ref{fig:geb_main}).

\section{Related Work}
In this section, we concisely situate our work with respect to literature on
level-sets, differentiable rendering and topological derivatives. A more
detailed discussion on literature relevant to our problem formulation, theory
and experimental settings can be found within Sections~\ref{sec:background},
\ref{sec:2d} and \ref{sec:3d}.

\paragraph{Level Sets} We use the level-set method~\cite{osher88} for shape evolution.
In the case of 3D, level-sets have been extensively used for shape
reconstruction from range data~\cite{whitaker1998level}, stereo
images~\cite{faugerasVariationalPrinciplesSurface},  multi-view
images~\cite{gargalloMinimizingReprojectionError2007} and RGB-D
data~\cite{newcombe2011kinectfusion}.
In 2D, they have been used for image segmentation as active contour
models~\cite{caselles1995geodesic,chan2001active,kass1988snakes}.
These methods derive equations of motion for curves and surfaces as a set of
partial differential equations that minimize an optimization objective.
Solem and Overgaard~\cite{solemGeometricFormulationGradient2005} provide a
geometric viewpoint to analyze these methods.
The first variational approach to level-set optimization was proposed by
Zhao~\etal~\cite{zhao1996variational} for multi-phase surface motion.
More recently, level-sets as signed-distance functions have become a popular
choice for generative modeling of 3D shapes~\cite{chen19,park19} and inverse
rendering~\cite{jiang20,liu20,niemeyer2020dvr,yariv20}.

\paragraph{Differentiable Rendering}
Differentiable renderers are designed to estimate the derivatives of a rendering
integral with the primary focus on handling geometric discontinuities.
One class of differentiable renderers handles primary-visibility discontinuity for triangle
meshes~\cite{laine20,liu19,loper2014opendr,ravi20} with a blurring kernel --- an
approach that is fast but not physically accurate.
Li~\etal~\cite{li18} propose edge sampling  to correctly handle discontinuities
and differentiate the full rendering
equation~\cite{kajiyaRENDERINGEQUATION1986}, which was later extended for
differentiable rasterization of vector
graphics~\cite{liDifferentiableVectorGraphics2020}.
Subsequently, several works aim for numerical efficiency and
accuracy~\cite{bangaru20,mitsuba3,loubet2019reparameterizing,zhang20}.
Work by Bangaru~\etal~\cite{bangaru2022differentiable} and
Vicini~\etal~\cite{vicini2022differentiable} is especially relevant to our work
as they handle discontinuities for surfaces represented by signed-distance
functions.
All such works measure image sensitivities with respect to visibility changes on
silhouette boundaries, while we focus on meaasuring them in the shape's
interior.

\paragraph{Topological Derivatives}
The standard definition of topological derivatives is given by Sokolowski and Zochowski~\cite{sokolowski1999topological}.
Part of our notation comes from the asymptoptic analysis of TDs in~\cite{nazarov2003asymptotic}.
They have been used in relatively low-dimensional
problems in physics for electrical impedence
tomography~\cite{hintermuller2008electrical} and inverse scattering~\cite{carpioTopologicalDerivativesShape2008}, or in structural engineering to discover optimal support structures~\cite{amstutz2010topological}.
In computer vision, the use of topological derivatives has been limited to a few problems in image processing~\cite{larrabide2008topological}.
The work by Burger~\etal~\cite{burgerIncorporatingTopologicalDerivatives2004} is perhaps the closest to ours in terms of using the level-set method with SDs and TDs.
Please refer to \cite{novotny2019applications} for a recent survey on the usage of TDs.

\section{Background and Problem Formulation}
\label{sec:background}
We consider the problem of inverse shape optimization from a set of images.
Our focus is on recovering closed curves in 2D and closed surfaces in 3D.
With $\surf$ being the surface under consideration, we take a variational
optimization approach to minimize an image functional,
\eqn{
    \label{eq:main}
    \min_\surf \energy(\surf) = \int_\image g(\pixel)\ d\pixel.
}
The functional $\energy$ is defined over an image plane $\image$ with $d\pixel$
as the pixel-area measure.
The integrand $g$ is evaluated at pixels $\pixel$ and can be any reasonable
error function.
For instance, $g = |\hat{L} - L|$ can evaluate the error between measured ($L$) and
estimated radiance $(\hat{L})$.

We define $\surf$ as a closed $n$-dimensional surface residing in $\R^{n+1}$,
where $n=1$ for planar curves and $n=2$ for surfaces in 3D.  
We use a level-set function $\phi:\R^{n+1}\to\R$ to represent $\surf$ as,
$\surf \coloneqq \{\x : \phi(\x) = 0 \}$.
We use the standard convention~\cite{osher88} to represent the interior of
$\surf$ as $\Omega_\surf = \{\x: \phi(\x) \leq 0\}$ and the exterior as
$\R^{n+1}\setminus\Omega_\surf$.

To recover the optimal surface, we require a notion of gradient in the
variational setting --- for which we build on the framework of variational
level-sets \cite{solemGeometricFormulationGradient2005}.
Consider the surface $\surf$ to be a point on an $m$-dimensional manifold $M$ of admissible surfaces.
The functional $\energy$ to be minimized can be considered a scalar function that maps $M\to\R$.
We use the idea of \textit{differentials} (and corresponding G\^ateaux derivatives) from differential geometry~\cite{milnor1997topology}.
We derive the gradient $\nabla_M\energy$ and optimize $\surf$ with an
initial value problem: $\dot\surf(t) = \nabla_M\energy(\surf(t))$,
with $\surf_0$ as the surface at initialization.
The optimal surface $\surf^*$ can be then be recovered using a regular surface
evolution of $t\mapsto\surf(t)$.
By definition, the gradient flow $\nabla\energy$ is composed of only normal components as
tangential flow fields keep $\surf$ invariant~\cite{osher88}.
We refer the reader to~\cite{solemGeometricFormulationGradient2005} for a more
rigorous discussion on why $\nabla I$ is a descent direction and reduces $I$.
Intuitively, $\nabla I$ is a scalar speed in the direction of the normal
$\n$ at each point $\x\in\surf$ such that $I$ is reduced.
The corresponding evolution equation for the underlying level-set function is,
\eqn{
    \frac{\partial\phi}{\partial t} &= -\nabla_M\energy |\nabla\phi(t)|.
    \label{eq:eul}
}
In Sections~\ref{sec:2d} and ~\ref{sec:3d} we derive evolution equations of this
form to minimize $I$ in the case of closed curves in 2D and surfaces in 3D
respectively.
Computing $\nabla\energy$ requires defining an inner product which operates on the
tangent space $T_\surf M$ at $\surf$.
For $w$ and $v$ normal velocity directions on $T_\surf M$, we can define
$\langle\cdot, \cdot\rangle_\surf : T_\surf M \times T_\surf M \to \R$ as,
\eqn{
    \langle w, v \rangle_\surf = \int_\surf w(\x)v(\x)\ d\sigma,
}
where $d\sigma$ is the surface measure.
Consider $v\in T_\surf M$ as any normal velocity direction of the form
$v=-\frac{\psi}{|\nabla\phi|}$.
We can deform the surface $\surf$ in this direction by updating the level-set
function as $\phi^s = \phi + s\psi$, where $s$ controls the level of
deformation and $\psi$ is the speed.
The directional (G\^ateaux) derivative of $\energy$ in this direction $v$ can then be defined
as~\cite{solemGeometricFormulationGradient2005},
\eqn{
    d\energy(\surf)v \coloneqq \frac{d}{ds} \energy(\phi + s\psi)\Big|_{s=0}.
    \label{eq:gat_def}
}
For \emph{any} such normal velocities $v$ in the tangent space, if we can
reformulate ($\ref{eq:gat_def}$) as,
\eqn{
    \label{eq:gat}
    d\energy(\surf)v = \frac{d}{ds} \energy(\phi + s\psi)\Big|_{s=0} \stackrel{?}{=} \langle
    w, v \rangle_\surf,
}
then $w$ is the gradient $\nabla_M\energy$.
We revisit this definition of G\^ateaux derivative (\ref{eq:gat}) and the
evolution equation (\ref{eq:eul}) in the subsequent sections to derive gradient
flows using shape and topological derivatives.

\section{Curves in 2D}
\label{sec:2d}
We derive equations of motion for closed planar curves subject to a rendering
functional.
These equations take the form of \textit{shape derivatives} (\S~\ref{sec:2d_sd}) and
\textit{topological derivatives} (\S~\ref{sec:2d_td}) and can be used to recover 2D vector shapes
with respect to an error function.
With this formulation, we show shape evolution  for
image vectorization (Figure~\ref{fig:2d_main}) and generation of vector graphics from text prompts (Figure~\ref{fig:sds}).
\subsection{Shape Derivatives}
\label{sec:2d_sd}

\res{
    \label{res:res1}
    Let $I$ be an image functional of a closed curve $\surf$ encoded as the
    0-isocontour of a level-set function $\phi$ in $\R^2$. If the interior of the
    shape corresponds to a foreground scene function $\fore$ and the exterior to a
    background function $\back$ then the shape derivative
    $\nabla I$ is,
    $$
        \nabla I = \fore - \back.
    $$
}
We start by looking at the image integral (\ref{eq:main}) that integrates a
scene function $g$ over the image plane,
\eqn{
    I(\surf) = \int_\image g(\pixel)\ d\pixel = \int_{\R^2} g(\x)\ d\x,
    \label{eq:domain_change}
}
where $g$ is any arbitrary function of color on the image plane.
For 2D rasterization, the image plane coincides with $\R^2$ and hence we can
change the integral domain as shown in (\ref{eq:domain_change}).
We also assume that $g$ is band-limited to the support of the image \ie,
$g(\pixel) = 0$ for $\pixel \notin \image$.
The closed curve $\surf$ partitions the $\R^2$ plane into two regions as per
Jordan's Curve Theorem~\cite{hales2007jordan}. We delineate the two regions with
scene functions $\fore$ and $\back$ for foreground and background respectively.
We use a level-set function $\phi$ to synthesize a characteristic function
$H~\circ~\phi$, where $H$ is a standard heaviside function. By construction, our
function evaluates to $0$ in the foreground and $1$ in the background.
The characteristic function can be used to expand the integrand as,
\begin{equation}
\normalsize
\qquad
\begin{aligned}
    \nonumber
I(\surf) = \int_{\R^2} \fore (\x) [1 - H(\phi(\x))] +\\ \back (\x)
H(\phi(\x))\ 
d\x.
\end{aligned}
\qquad
\footnotesize
\vcenter{\hbox{\begingroup%
  \makeatletter%
  \providecommand\color[2][]{%
    \errmessage{(Inkscape) Color is used for the text in Inkscape, but the package 'color.sty' is not loaded}%
    \renewcommand\color[2][]{}%
  }%
  \providecommand\transparent[1]{%
    \errmessage{(Inkscape) Transparency is used (non-zero) for the text in Inkscape, but the package 'transparent.sty' is not loaded}%
    \renewcommand\transparent[1]{}%
  }%
  \providecommand\rotatebox[2]{#2}%
  \newcommand*\fsize{\dimexpr\f@size pt\relax}%
  \newcommand*\lineheight[1]{\fontsize{\fsize}{#1\fsize}\selectfont}%
  \ifx\svgwidth\undefined%
    \setlength{\unitlength}{54.90992208bp}%
    \ifx\svgscale\undefined%
      \relax%
    \else%
      \setlength{\unitlength}{\unitlength * \real{\svgscale}}%
    \fi%
  \else%
    \setlength{\unitlength}{\svgwidth}%
  \fi%
  \global\let\svgwidth\undefined%
  \global\let\svgscale\undefined%
  \makeatother%
  \begin{picture}(1,0.99574527)%
    \lineheight{1}%
    \setlength\tabcolsep{0pt}%
    \put(0,0){\includegraphics[width=\unitlength,page=1]{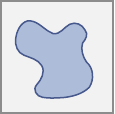}}%
    \put(0.76843309,0.82764607){\makebox(0,0)[lt]{\lineheight{1.25}\smash{\begin{tabular}[t]{l}$\back$\end{tabular}}}}%
    \put(0.36744573,0.80296914){\makebox(0,0)[lt]{\lineheight{1.25}\smash{\begin{tabular}[t]{l}$\surf$\end{tabular}}}}%
    \put(0.36378205,0.21061399){\makebox(0,0)[lt]{\lineheight{1.25}\smash{\begin{tabular}[t]{l}$\fore$\end{tabular}}}}%
    \put(0.05096525,0.08516277){\makebox(0,0)[lt]{\lineheight{1.25}\smash{\begin{tabular}[t]{l}$\image$\end{tabular}}}}%
    \put(0.15621231,0.68954111){\rotatebox{-35.862818}{\makebox(0,0)[lt]{\lineheight{1.25}\smash{\begin{tabular}[t]{l}$H\circ\phi =0$\end{tabular}}}}}%
    \put(0,0){\includegraphics[width=\unitlength,page=2]{fig17.pdf}}%
  \end{picture}%
\endgroup%
}}
\normalsize
\normalsize
\end{equation}

We can now the derive the shape derivative $\nabla I$ using a perturbation of the
form $\phi^s = \phi + s\psi$ that corresponds to velocity $v$.
The G\^ateaux derivative of $I$ in this direction $v$ is,
\begin{align}
            dI(\surf)v &= \frac{d}{ds} I(\phi + s\psi)\Big|_{s=0} \qquad\lhd\text{From
            (\ref{eq:gat_def})} \nonumber \\
                    &= \int_{\R^2} \fore \frac{d}{ds}[1 - H(\phi^s)] +
                    \back  \frac{d}{ds}H(\phi^s)\ d\x\ \Big|_{s=0}\
                    \nonumber\\
                    &= \int_{\R^2} -\delta(\phi)\psi\fore +
                    \delta(\phi)\psi\back\ d\x \nonumber \\
                    &= \int_{\R^2} (\fore -
                    \back)\frac{-\psi}{|\nabla\phi|}\delta(\phi)|\nabla\phi|
                    \ d\x \nonumber \\
                    &= \int_\surf (\fore - \back) v\ d\sigma
                    = \langle \fore - \back, v \rangle_\surf,
    \label{eq:2d_derivation}
\end{align}
where $d\sigma = \delta(\phi(\x))|\nabla\phi(\x)|d\x$ is the surface measure \cite{hormander2015analysis,solemGeometricFormulationGradient2005} and $v =
\frac{-\psi}{|\nabla\phi|}$. Note the interchange of integral and differential
operators due to Leibniz rule. Comparing the definitions in (\ref{eq:gat}) and (\ref{eq:2d_derivation}), we conclude that the shape derivative for functional
$I$ is,
\eqn{
    \nabla I = \fore - \back.
    \label{eq:2d_sd}
}
The shape derivative in (\ref{eq:2d_sd}) can be used to minimize $I$ using the
level-set evolution equation (\ref{eq:eul}).
As per the domain of the integral in (\ref{eq:2d_derivation}), this derivative
is defined only on the curve $\surf$.
Li~\etal~\cite{liDifferentiableVectorGraphics2020} arrive at a similar result
using differential calculus and numerically evolve 2D shapes by explicitly
sampling the edges.
As shown in Figure~\ref{fig:2d_main}, SDs can fail to prompt crucial topology
changes for image vectorization.

\begin{figure}
\begin{minipage}[c]{\linewidth}
    \centering
    \footnotesize
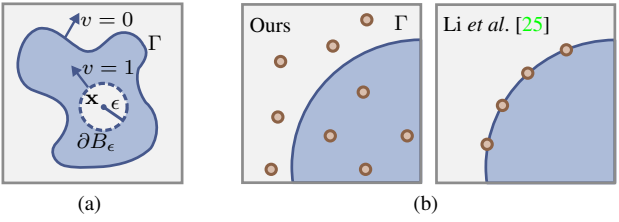
\end{minipage}
\caption{
    (a) We derive topological derivatives for 2D rendering functionals. An
    infinitesimal-hole of perturbation $B_\epsilon$ with radius $\epsilon\to 0$ is
    introduced at point $\x$. A G\^ateaux derivative is computed in the
    direction of a normal velocity $v$ such that $v = 1$ on $\partial
    B_\epsilon$ and $0$ on the unperturbed surface $\surf$.
    (b) We find functional forms of shape and topological
    derivatives to be the same with the only difference being the domain.
    Numerically, this eliminates the need for explicitly sampling boundaries for
    shape evolution as in \cite{liDifferentiableVectorGraphics2020}.
}
    \label{fig:2d_construction}
\end{figure}

\subsection{Topological Derivatives}
\label{sec:2d_td}
The topological derivative measures sensitivities with respect to
infinitesimally small perturbations in a shape's volume.
We derive the topological derivative in the context of inverse rendering to
inform where to nucleate holes and phases in a shape in order to reduce a
rendering functional.
\res{
    \label{res:res2}
    Consider a closed curve $\surf$ with $\Omega_\surf$ as its interior. For a
    point $\x\in\R^2$
    the topological derivative with respect to hole and phase nucleation
    is,
    $$
    D_\tau(\x) = \fore(\x) - \back(\x).
    $$
}
We follow the standard definition of a topological
derivative from Sokolowski and Zochowski~\cite{sokolowski1999topological}.
We focus on hole nucleation in this section and defer the derivation for phase
nucleation to the Supplementary.
Let $\Omega_\surf$ denote the interior of a given curve $\surf$.
For a point $\pixel\in\image$ on the image plane, we introduce a small circular
hole $B_\epsilon(\pixel)$ of radius $\epsilon$.
The topological derivative of the image functional $I$ at a point
$\x\in\Omega_\surf$ can then be defined as a scalar function,
\eqn{
    D_\tau(\x, \Omega_\surf) \coloneqq \lim_{\epsilon\to 0} \frac{I(\Omega_\surf\setminus
    \overbar{B_\epsilon(\x)}) - I(\Omega_\surf)}{V(B_\epsilon)},
    \label{eq:td_define}
}
where $I(\Omega_\surf)$ is the functional value with the unperturbed shape,
$\overbar{B_\epsilon}$ is the closure set of points inside the perturbation and
$V(B_\epsilon) = \pi\epsilon^2$ is the area of the circle.
The corresponding asymptotic expansion of this
definition~\cite{nazarov2003asymptotic,sokolowski1999topological} is,
\eqn{
    I(\Omega_\surf\setminus\overbar{B_\epsilon}) = I(\Omega_\surf) + V(B_\epsilon)D_\tau(\x,
    \Omega_\surf) + o(V(B_\epsilon)).
    \label{eq:asymptotic}
} \normalsize
We compute the G\^ateaux derivative of this expression in the direction of a
normal velocity $v$ defined as shown in Figure~\ref{fig:2d_construction} (a).
Intuitively, the defined velocity increases the size of the perturbation by a
constant and keeps the curve $\surf$ as it is.
Formally, $v=0$ on the unperturbed curve $\surf$ and $v=1$ at the boundary
$\partial B_\epsilon$ of the perturbation hole, such that it smoothly goes to
$0$ outside a small neighborhood of $\x$.
Assuming $\x\notin\surf$ and since velocity $v$ is $0$ on the unperturbed curve,
the shape derivative (\ref{eq:2d_sd}) in the direction $v$ is $0$:

\eqn{
    dI(\Omega_\surf) v = \int_\surf (\fore - \back ) v\ d\sigma = 0.
    \label{eq:directional_omega}
}
From (\ref{eq:td_define}), (\ref{eq:asymptotic}) and (\ref{eq:directional_omega})  we can rewrite the TD as:
\eqn{
    \spl{
        D_\tau(\x, \Omega_\surf) &= \lim_{\epsilon\to
        0}\frac{1}{V'(B_\epsilon)}dI(\Omega_\surf\setminus
        \overbar{B_\epsilon})v\\
                                 &= \lim_{\epsilon\to
                                     0}
                                     \frac{1}{V'(B_\epsilon)}\int_{\surf\cup\partial
                                     B_\epsilon}
                                     (\fore - \back ) v\ d\sigma.
    }
    \label{eq:2d_td_rework}
}
\begin{figure}
\begin{minipage}[c]{\linewidth}
    \centering
    \footnotesize
\begingroup%
  \makeatletter%
  \providecommand\color[2][]{%
    \errmessage{(Inkscape) Color is used for the text in Inkscape, but the package 'color.sty' is not loaded}%
    \renewcommand\color[2][]{}%
  }%
  \providecommand\transparent[1]{%
    \errmessage{(Inkscape) Transparency is used (non-zero) for the text in Inkscape, but the package 'transparent.sty' is not loaded}%
    \renewcommand\transparent[1]{}%
  }%
  \providecommand\rotatebox[2]{#2}%
  \newcommand*\fsize{\dimexpr\f@size pt\relax}%
  \newcommand*\lineheight[1]{\fontsize{\fsize}{#1\fsize}\selectfont}%
  \ifx\svgwidth\undefined%
    \setlength{\unitlength}{231.76800728bp}%
    \ifx\svgscale\undefined%
      \relax%
    \else%
      \setlength{\unitlength}{\unitlength * \real{\svgscale}}%
    \fi%
  \else%
    \setlength{\unitlength}{\svgwidth}%
  \fi%
  \global\let\svgwidth\undefined%
  \global\let\svgscale\undefined%
  \makeatother%
  \begin{picture}(1,0.31065547)%
    \lineheight{1}%
    \setlength\tabcolsep{0pt}%
    \put(0,0){\includegraphics[width=\unitlength,page=1]{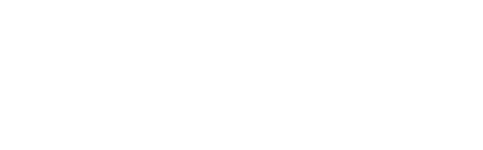}}%
    \put(0.04403953,0.0288055){\makebox(0,0)[lt]{\lineheight{1.25}\smash{\begin{tabular}[t]{l}Raster Input\end{tabular}}}}%
    \put(0,0){\includegraphics[width=\unitlength,page=2]{fig7.pdf}}%
    \put(0.27982641,0.03514564){\makebox(0,0)[lt]{\lineheight{1.25}\smash{\begin{tabular}[t]{l}TD\end{tabular}}}}%
    \put(0,0){\includegraphics[width=\unitlength,page=3]{fig7.pdf}}%
    \put(0.27987997,0.17869822){\makebox(0,0)[lt]{\lineheight{1.25}\smash{\begin{tabular}[t]{l}SD~\cite{liDifferentiableVectorGraphics2020}\end{tabular}}}}%
    \put(0,0){\includegraphics[width=\unitlength,page=4]{fig7.pdf}}%
  \end{picture}%
\endgroup%

\end{minipage}
    \caption{\textbf{Reconstruction of vector images from raster input.}
    Shape derivatives (analagous to curve gradients
    in~\cite{liDifferentiableVectorGraphics2020}) evolve closed curves with
    visibility changes only on the boundaries.
    These gradients are sparse and the optimization can plateau at a local
    minimum (\emph{top}).
    Topological derivatives can reconstruct a vector image
    (\emph{bottom}) from a raster input (\emph{left}) with adaptive topological
    changes.
    }
\label{fig:2d_main}
\end{figure}

Note that with the introduction of a hole, the perturbed shape is comprised of
surface elements from the original domain $\surf$ in addition to the boundary
$\partial B_\epsilon$ of the hole --- that is, $\hat{\surf} = \surf \cup \partial
B_\epsilon$ where $\hat{\surf}$ is the perturbed curve (see
Figure~\ref{fig:2d_construction} (a)).
The denominator of (\ref{eq:2d_td_rework}) is a result of the fact that
$dV(\epsilon)v = V'(\epsilon) = 2\pi\epsilon$ as $v=1$ on the hole's boundary.
Further simplification yields:
\eqn{
    \spl{
        D_\tau(\x, \Omega_\surf) &=  \lim_{\epsilon\to
                                     0}
                                     \frac{1}{2\pi\epsilon}\int_{\surf} (\fore -
                                     \back)v\ d\sigma+\\
                                     &\hspace{6em} \int_{\partial
                                     B_\epsilon}
                                     (\fore - \back )v\ d\sigma\\
        &=\lim_{\epsilon\to
                                     0}
                                     \frac{1}{2\pi\epsilon}\int_{\partial
                                     B_\epsilon}
                                     (\fore - \back )\ d\sigma\\
                                 &= \fore(\x) - \back (\x).
    }
    \label{eq:2d_td}
}
Equation~\ref{eq:2d_td} is the topological derivative of the image functional
$I$ at a point $\x$.
Intuitively, at a given point in the interior of the shape, if the error with
respect to the background color is lower than the foreground, formation of a
hole can be prompted by increasing the value of $\phi(\x)$. 
We note that $D_\tau$ is defined in $\R^2\setminus\surf$.
This is in stark contrast with the  shape derivatives from Result 1 that encouraged visibility changes only on the curve.
By analogy, we also derive topological derivatives for phase nucleation and
arrive at the same result, \ie $D_\tau(\x, \R^2\setminus\Omega_\surf) =
\fore(\x) - \back(\x)$, where $\x$ is a point in the exterior of the given
shape.
The full derivation is in the Supplementary.

\paragraph{Level-Set Evolution}
We can now formulate the evolution equation in (\ref{eq:eul}) using both shape and topological derivatives:
\eqn{
    \frac{\partial\phi}{\partial t} = -\left[\nabla I(\x) +
    D_\tau(\x)\right]|\nabla\phi|.
    \label{eq:2d_evolution_1}
}
The SD $\nabla I$ is defined on $\surf$ and is $0$ elsewhere.
\begin{figure}
\begin{minipage}[c]{\textwidth}
    \small
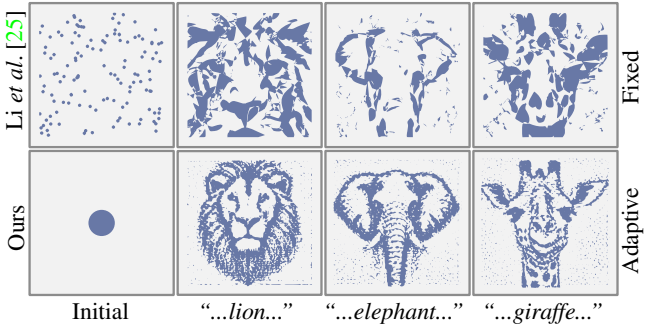
\end{minipage}
    \caption[Caption for LOF]{\textbf{Adaptive generation of detailed vector graphics.} 
    Our formulation for surface evolution can be used to induce
    topological changes subject to any differentiable loss functions for images.
    Here we show an example of generating vector graphics from a text prompt
    using the score-distillation-sampling loss from~\cite{poole2022dreamfusion}.
    (\emph{top}) Existing methods use differentiable vectorization
    from~\cite{liDifferentiableVectorGraphics2020} and are forced to work with
    fixed shape topologies.
    (\emph{bottom}) Conversely, topological derivatives can be used to
    adaptively generate detailed
    vector graphics from a simple initialization (\emph{bottom-left}).
    \setcounter{footnote}{1}
    \footnotemark[1]
    }
\label{fig:sds}
\end{figure}
\footnotetext[1]{Text-prompt: ``Frontal face of \texttt{animal}. Minimal line drawing.
Trending on artstation. Plain white background. Black and white.''}

The TD $D_\tau$ is defined on $\R^2\setminus\surf$.
As shown in Figure~\ref{fig:2d_main}, we can use the evolution equation for
image-based reconstruction of vector graphics from raster input.
When only the first term $\nabla I$ is used, the optimization can plateau at a
local minimum.
The second term is critical for recovering the target geometry, particularly
when the target has a differing geometric structure from the initialization.
We show more results in this context along with the implementation details in the
Supplementary.

\paragraph{Eliminating Edge Sampling}
Comparing Result 1 and Result 2, we find that the topological and shape
derivatives are exactly the same except for the domain on which they are
defined.
This simplifies the evolution equation in (\ref{eq:2d_evolution_1}) with a
single term for the entirety of $\R^2$,
\eqn{
    \frac{\partial\phi}{\partial t} = (\back - \fore)|\nabla\phi|.
}
In practice, for numerical optimization, we find that this simplification
eliminates the need for explicitly sampling boundaries for shape
evolution as done in Li~\etal's~\cite{liDifferentiableVectorGraphics2020}
method for differentiable vector graphics (Figure~\ref{fig:2d_construction} (b)).
For complex shapes, with a large set of paths, this leads to faster
optimization. We provide details in the Supplementary.

\paragraph{Generative Vector Graphics}
Using the chain-rule, the evolution equation can be adapted for problems beyond
just reconstruction.
Given an aribitrary differentiable loss function $\mathcal{L}$, we
can evolve the shapes with topological changes such that $\mathcal{L}$ is
minimized by the following evolution:
\eqn{
    \frac{\partial\phi}{\partial t} = \frac{\partial\mathcal{L}}{\partial
    I}(\back - \fore)|\nabla\phi|.
}
We show an application of generating 2D vector graphics from text-prompts with
a text-to-image diffusion model~\cite{rombach2022high}.
The loss function $\mathcal{L}$ is the score-distillation-sampling loss
from~\cite{poole2022dreamfusion}.
Previous methods for text-to-svg use the differentiable rasterizer
from~\cite{liDifferentiableVectorGraphics2020}.
Such methods require a fixed number of shapes starting from the intialization
and cannot adaptively change the topology in a differentiable manner.
Using topological derivatives, we show highly-detailed generation of vector
graphics from a relatively simple initialization of a disk (Figure~\ref{fig:sds}).
More examples and implementation details are in the Supplementary.

\section{Surfaces in 3D}
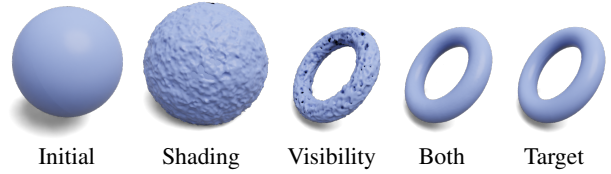
\begin{figure}
\begin{minipage}[c]{\textwidth}
    \small
\begingroup%
  \makeatletter%
  \providecommand\color[2][]{%
    \errmessage{(Inkscape) Color is used for the text in Inkscape, but the package 'color.sty' is not loaded}%
    \renewcommand\color[2][]{}%
  }%
  \providecommand\transparent[1]{%
    \errmessage{(Inkscape) Transparency is used (non-zero) for the text in Inkscape, but the package 'transparent.sty' is not loaded}%
    \renewcommand\transparent[1]{}%
  }%
  \providecommand\rotatebox[2]{#2}%
  \newcommand*\fsize{\dimexpr\f@size pt\relax}%
  \newcommand*\lineheight[1]{\fontsize{\fsize}{#1\fsize}\selectfont}%
  \ifx\svgwidth\undefined%
    \setlength{\unitlength}{236.29607391bp}%
    \ifx\svgscale\undefined%
      \relax%
    \else%
      \setlength{\unitlength}{\unitlength * \real{\svgscale}}%
    \fi%
  \else%
    \setlength{\unitlength}{\svgwidth}%
  \fi%
  \global\let\svgwidth\undefined%
  \global\let\svgscale\undefined%
  \makeatother%
  \begin{picture}(1,0.28946735)%
    \lineheight{1}%
    \setlength\tabcolsep{0pt}%
    \put(0,0){\includegraphics[width=\unitlength,page=1]{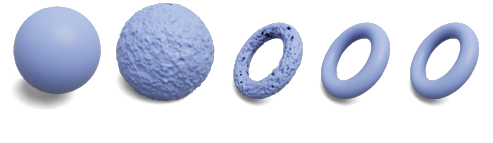}}%
    \put(0.06824665,0.02403792){\color[rgb]{0,0,0}\makebox(0,0)[lt]{\lineheight{1.25}\smash{\begin{tabular}[t]{l}Initial\end{tabular}}}}%
    \put(0.26544598,0.02403792){\color[rgb]{0,0,0}\makebox(0,0)[lt]{\lineheight{1.25}\smash{\begin{tabular}[t]{l}Shading\end{tabular}}}}%
    \put(0.46572751,0.02403792){\color[rgb]{0,0,0}\makebox(0,0)[lt]{\lineheight{1.25}\smash{\begin{tabular}[t]{l}Visibility\end{tabular}}}}%
    \put(0.67486772,0.02403792){\color[rgb]{0,0,0}\makebox(0,0)[lt]{\lineheight{1.25}\smash{\begin{tabular}[t]{l}Both\end{tabular}}}}%
    \put(0.84303738,0.02403792){\color[rgb]{0,0,0}\makebox(0,0)[lt]{\lineheight{1.25}\smash{\begin{tabular}[t]{l}Target\end{tabular}}}}%
  \end{picture}%
\endgroup%

\end{minipage}
\caption{\textbf{Visibility gradients drive topological changes.} Given a set of
    multi-view images, we recover a genus-1 shape from a sphere. When shading
    gradients are used the optimization makes only local updates. Using
    visibility gradients the topology and the silhouettes of the recovery match
    the target.  Both the terms are required for optimal recovery.}
    \label{fig:vis_shade_gradients}
\end{figure}

\label{sec:3d}
Similar to the 2D case, our goal is to minimize the image functional $I$ defined
in (\ref{eq:main}). 
Comparing it to the rendering equation~\cite{kajiyaRENDERINGEQUATION1986}, $I$
can depend on scene parameters such as the geometry, material and lighting in
the scene.
We focus on geometry optimization and take the material and lighting parameters
as given.
The function $g$ is discontinuous in terms of the geometry parameters and hence
naively differentiating $I$ using automatic differentiation is
erroneous~\cite{li18}.

\subsection{Background on Shading and Visibility}
\label{sec:3d_sd}
To address discontinuities with respect to visibility changes,
Gargallo~\etal~\cite{gargalloMinimizingReprojectionError2007} derive the
shape derivative of $I$ to evolve a level-set function $\phi$ as, %
\eqn{
    dI (\surf) = -\underbrace{\nabla
    g\cdot\frac{\x}{\x^3_z}}_\textrm{Shading}\visibility + \underbrace{(g -
    g_B)\frac{\x\cdot\nabla(\n\cdot\x)}{\x^3_z}\delta(\n\cdot\x)}_\textrm{Visibility}\visibility,
    \label{eq:differential_3d}
}
where $\visibility$ denotes the visibility, $\n$ the normal and $\x_z$ the $z$ coordinate.
The first term here is similar to shading gradients in more recent
differentiable renderers~\cite{mitsuba3,laine20,li18,ravi20}, which can be used
to make local updates in the interior of the visible surface.
\begin{figure}
\begin{minipage}[c]{\linewidth}
    \small
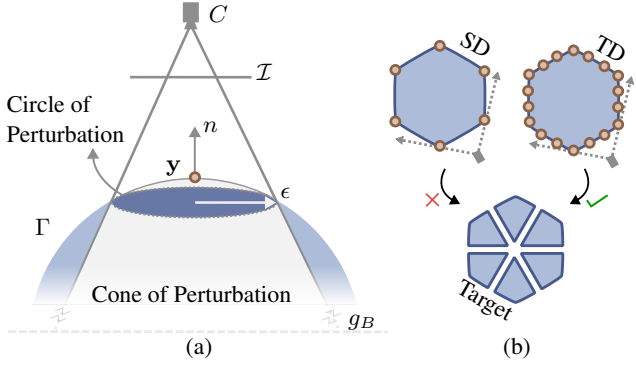
\end{minipage}
    \caption{(a) We derive topological derivative at a point $\y$ on a surface
    $\surf$ using a vanishing cone of perturbation originating at the camera
    $C$. (b) Visibility terms from shape derivatives (SD) are evaluated
    only at the apparent contours. In flatland, these points resemble the extreme
    points of the convex set of the shape. A target shape with the same convex
    booundary as the initial shape cannot be recovered using the visibility terms in
    SDs. TDs evaluate visibility changes on the entire surface and can resolve
    such ambiguities.}
\label{fig:td_construction}
\end{figure}

The second term is enabled only at the silhouette boundary as per $\delta(\n\cdot\x)$,
where $\delta$ is a dirac distribution function.
The difference $g - \back$ moves the visible contours of the surface based on
the foreground ($g$) and background ($g_B$) radiance.
This visibility term is similar to shape derivatives in the 2D case
(\ref{eq:2d_sd}), with additional factors to account for the distance from
the camera, visibility and shape curvature.

\paragraph{Need for Topological Derivatives} 
For recovering geometry with complex topology, we find the
visibility gradients to be critical in terms of recovering the overall
structure.
We empirically show this in Figure~\ref{fig:vis_shade_gradients}, where a
genus-1 shape is optimized from a sphere using the two gradient forms.
When only shading gradients are used, the optimization fails to evolve the
topology; with visibility gradients in isolation we can retrieve the correct
topology but not the details; when both are used together, we recover the target shape.
Despite this significance, the
optimization signal obtained from visibility gradients is quite sparse.
As evident from (\ref{eq:differential_3d}), these gradients are evaluated only
at the extremeties of the shape and cannot induce visibility changes in the
interior of the shape.
In flatland, this means that visibility gradients account for changes
only at the extreme points of the convex set covering a shape (as
shown in Figure~\ref{fig:td_construction} (b)).
This makes the optimization susceptible to unexpected local minima as
topological changes are less likely in regions away from the surface.
Our goal is to resolve this by enabling topological changes in the interior
regions using gradients that measure visibility changes on the \textit{entire}
surface.
\subsection{Topological Derivatives}
\label{sec:3d_td}
\res{
    \label{res:res3}
    Let $I$ be an image functional of a closed and connected surface $\surf$ in
    $\R^3$. 
    The functional $I$ integrates a scene function $g$ for the
    surface and $g_B$ for the background. Then the topological derivative of $I$ at a
    point $\y\in\surf$ with respect to an infinitesimal conical perturbation
    from the origin through $\y$ is,
    $$
    D_\tau(\y, \surf) = (g(\y) - g_B(\y))\frac{\kappa\y^t\y}{\y^3_z}.
    $$
}
In case of surfaces in $\R^3$, there is a large class of infinitesimal
perturbations that change the configuration of the shape.
In this work we focus on perturbations that influence the genus of the shape.
We begin by considering a point $\y$ on a surface $\surf$ at which want to
estimate the topological derivative.
Without loss of generality, we will assume that the surface $\surf$ is placed
in between the camera at the origin and a background scene.
A curved circle of planar radius $\epsilon$ is placed on the surface with $\y$
as the center (Figure~\ref{fig:td_construction} (a)).
We term the corresponding planar circle as the circle of perturbation.
Starting from the origin, we can find an elliptic cone that inscribes this
circle at an angle that is consistent with the normal at $\y$. 
We use this cone to define our perturbed shape as the difference of the
enclosed volume of the original shape and the intersection of the original shape
with the cone of perturbation.
If $\Omega_\surf$ is the interior of $\surf$ and $\Omega_\epsilon$ of the
perturbation cone, then $\Omega_{\hat{\surf}(\epsilon)} = \Omega_\surf -
\Omega_\surf \cap\Omega_\epsilon$
is the volume enclosed by the perturbed surface $\hat{\surf}$.
By construction, we know that as $\epsilon \to 0$, the perturbed surface
$\hat{\surf}(\epsilon) \to \surf$.
Beyond just the perturbation, we are interested in its resultant effect on the
image.
The image plane intersects with the cone of perturbation and inscribes an ellipse.
Intuitively, as we increase the radius of the circle of perturbation, the area
of the ellipse on the image plane increases and the image accumulates more light
from the background.

In the given construction, we now define the topological derivative of $I$ at
$\y$ as,
\eqn{
    D_\tau(\y, \surf) \coloneqq \lim_{\epsilon\to 0}\frac{I(\hat{\surf})
    - I(\surf)}{V(\epsilon)}.
    \label{eq:td_3d_def}
}
Note that this is a slightly different notion of the topological derivative as
compared to the standard form~\cite{sokolowski1999topological} which introduces
vanishing balls $B_\epsilon$ at the points of perturbation.
Our deviation from this norm is out of necessity.
The error functionals that we consider for inverse rendering operate only on
the projection of the shape and hence any perturbations of the standard form in
the shape's interior will have no effect on the error.
The term $V(\epsilon)$ in the standard form is the Lebesgue measure of the ball
$B_\epsilon$, although, in our case, we choose $V(\epsilon) = \pi\epsilon^2$ as
the area of the planar circle.
This choice will lend us a limit (\ref{eq:td_3d_def}) that exists and is
finite.
\begin{figure}
\begin{minipage}[c]{\textwidth}
    \footnotesize
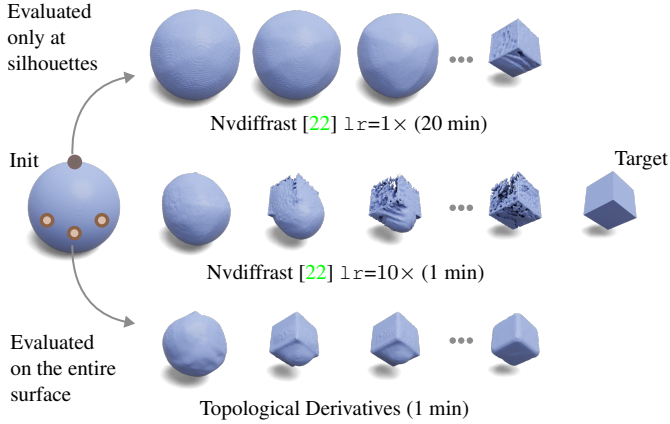
\end{minipage}
    \caption{\textbf{Large steps in visibility optimization.}  Standard
    differentiable renderers evaluate visibility gradients only at the
    silhouette.  Topological derivatives estimate sensitivies with respect to
    visibility changes on the entire surface. This enables a more robust way to
    take large steps in inverse rendering. We compare topological derivatives
    (\emph{bottom}) with visibility gradients (\emph{top}) from Nvdiffrast~\cite{laine20} for
    multi-view reconstruction. (\emph{middle}) We observe that silhouette gradients are unstable at
    higher learning rate (\texttt{lr}).
    }
\label{fig:large_steps}
\end{figure}

\setlength\intextsep{0pt}
\setlength{\columnsep}{5pt}%
\begin{wrapfigure}{r}{0pt}
    \footnotesize
\begingroup%
  \makeatletter%
  \providecommand\color[2][]{%
    \errmessage{(Inkscape) Color is used for the text in Inkscape, but the package 'color.sty' is not loaded}%
    \renewcommand\color[2][]{}%
  }%
  \providecommand\transparent[1]{%
    \errmessage{(Inkscape) Transparency is used (non-zero) for the text in Inkscape, but the package 'transparent.sty' is not loaded}%
    \renewcommand\transparent[1]{}%
  }%
  \providecommand\rotatebox[2]{#2}%
  \newcommand*\fsize{\dimexpr\f@size pt\relax}%
  \newcommand*\lineheight[1]{\fontsize{\fsize}{#1\fsize}\selectfont}%
  \ifx\svgwidth\undefined%
    \setlength{\unitlength}{77.55590724bp}%
    \ifx\svgscale\undefined%
      \relax%
    \else%
      \setlength{\unitlength}{\unitlength * \real{\svgscale}}%
    \fi%
  \else%
    \setlength{\unitlength}{\svgwidth}%
  \fi%
  \global\let\svgwidth\undefined%
  \global\let\svgscale\undefined%
  \makeatother%
  \begin{picture}(1,0.78812132)%
    \lineheight{1}%
    \setlength\tabcolsep{0pt}%
    \put(0,0){\includegraphics[width=\unitlength,page=1]{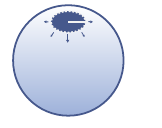}}%
    \put(0.0360256,0.0816746){\makebox(0,0)[lt]{\lineheight{1.25}\smash{\begin{tabular}[t]{l}$\surf$\end{tabular}}}}%
    \put(0.36413867,0.45783658){\makebox(0,0)[lt]{\lineheight{1.25}\smash{\begin{tabular}[t]{l}$v=1$\end{tabular}}}}%
    \put(0.64826812,0.69219053){\makebox(0,0)[lt]{\lineheight{1.25}\smash{\begin{tabular}[t]{l}$v=0$\end{tabular}}}}%
    \put(0,0){\includegraphics[width=\unitlength,page=2]{fig14.pdf}}%
  \end{picture}%
\endgroup%

\end{wrapfigure}

\begin{figure*}
\begin{minipage}[c]{\textwidth}
    \small
    \centering
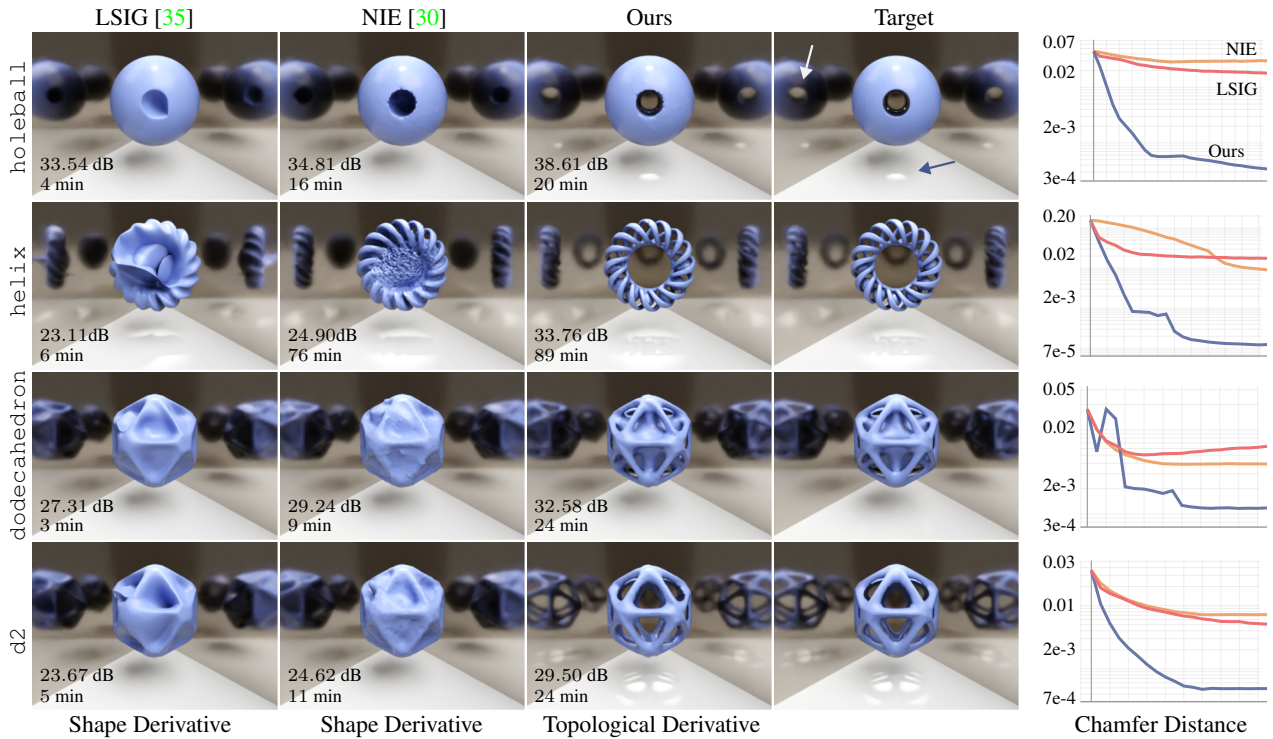
\end{minipage}
    \caption{\textbf{Inverse rendering of geometry with complex topology.} Given
    a set of images, we cover complex shapes from a spherical initialization.
    We compare our method that uses topological derivatives to a differentiable
    renderer~\cite{laine20} that uses visibility gradients defined on the
    silhouette. We regularize the gradients from~\cite{laine20} with two recent
    methods. LSIG~\cite{laine20} uses a triangle mesh representation and can
    make topological changes using remeshing~\cite{botsch2004remeshing}.
    NIE~\cite{mehta2022level} uses the same gradients with a level-set
    function.  These approaches struggle with prompting visibility changes in
    the  interior and can get stuck in a local minimum. Convergence plots of
    chamfer distance are on the right. We also provide PSNR and optimization
    time for each scene.  Our method, as expected, requires a slight
    time-overhead over NIE to estimate the TD. We place the recovered shapes in
    front of two glossy surfaces to highlight the quality of recovery.
    }
\label{fig:inverse_main}
\end{figure*}

Similar to the 2D case (\S~\ref{sec:2d_td}), we define a normal velocity $v$
such that $v=1$ on the newly formed visible contour around the circle of
perturbation, and $v=0$ on the unperturbed surface.
Hence, the G\^ateaux derivative of the image functional $I(\surf)$ in the direction $v$ is,
\eqn{
    \langle dI(\surf),v\rangle_\surf = \int_\surf dI(\surf)v\ d\sigma =
    0,
}
where $dI$ is defined as in (\ref{eq:differential_3d}).
Analogous to the case of TDs in 2D (\ref{eq:asymptotic}), by using asymptotic
expansion we can rework the definition for 3D
in (\ref{eq:td_3d_def}) as,
\eqn{
    D_\tau(\y, \surf) = \lim_{\epsilon\to
    0}\frac{1}{V'(\epsilon)}dI(\hat{\surf})v.
    \label{eq:td_3d_new_def}
}
Intuitively, as $\epsilon\to0$, this definition measures the rate of change in the image of a
perturbed shape with respect to the rate of increase in the area of the circle of
perturbation.
Using the shape derivative defined in (\ref{eq:differential_3d}) we can
replace the $dI(\hat{\surf})$ with,
\eqn{
    \nonumber
    D_\tau(\y, \surf) = &\lim_{\epsilon\to 0}\frac{1}{V'(\epsilon)} \int_{\hat{\surf}}\left[-\nabla
    g\cdot\frac{\x}{\x^3_z}\visibility\right]v\ +\\
    &\left[(g -
    g_B)\frac{\x\cdot\nabla(\n\cdot\x)}{\x^3_z}\delta(\n\cdot\x)\visibility\right]v\ d\sigma.
}
We simplify this integral with two assumptions. First, in the small neighborhood
of $\y$, as $\epsilon \to 0$, the radiance function $g$ is constant. This
eliminates the first term in the integrand as $\nabla g = 0$.
Next, we assume constant curvature. This results in
$\x\cdot\nabla(\n\cdot\x)=\kappa \x^t\x$ and curvature
\footnote{More details on the relevance
of $\kappa$ are in the Supplementary.}
$\kappa$ can be taken out of the integral sign
as, 
\eqn{
    \!\!\!\! D_\tau(\y, \surf) = \lim_{\epsilon\to
    0}\frac{\kappa}{V'(\epsilon)}\int_{\hat{\surf}} (g -
    g_B)\frac{\x^t\x}{\x^3_z}\delta(\n\cdot\x)\visibility v d\sigma.
}
The term $\delta (\n\cdot\x)\visibility$ constrains the domain for the integral
to the set of apparent contours that are visible from the camera.
The choice of $v$ further restricts this domain to only the circle of
perturbation (which we denote by $\partial\circ$) as by definition, $v=0$ for
all other contours:
\eqn{
    \nonumber
    D_\tau(\y, \surf) &= \lim_{\epsilon\to
    0}\frac{\kappa}{V'(\epsilon)}\int_{\partial\circ} (g -
    g_B)\frac{\x^t\x}{\x^3_z}\ d\sigma\\
    &=(g(\y) - g_B(\y))\frac{\kappa\y^t\y}{\y^3_z}.
    \label{eq:td_3d_final}
}
This is the topological derivative of functional $I$ at a visible point $\y$ on
the surface.
Unlike the visibility gradients by
Gargallo~\etal~\cite{gargalloMinimizingReprojectionError2007} in
(\ref{eq:differential_3d}) and other differentiable
renderers~\cite{mitsuba3,laine20,li18,ravi20}, the topological derivative in
(\ref{eq:td_3d_final}) can
prompt visibility changes on the entire surface.

\paragraph{Level-Set Evolution}
We can finally use topological derivatives (\ref{eq:td_3d_final}) and shape
derivatives (\ref{eq:differential_3d}) together to evolve a level-set function
$\phi$ using the evolution equation from (\ref{eq:eul}): 
\eqn{
    \frac{\partial\phi}{\partial t} = -dI(\surf)|\nabla\phi| - D_\tau(\x,
    \surf)|\nabla\phi|.  
}
Although, the shape derivative term ($dI(\surf)$) is valid only for primary
visibility.
We can lift this assumption by replacing the shape derivatives with gradients
from a differentiable path tracer for triangle meshes~\cite{mitsuba3,li18} as
follows~\cite{mehta2022level, remelli20},
\eqn{
    \frac{\partial\phi}{\partial t} = -\frac{\partial I}{\partial\x}\cdot\n -
    D_\tau(\x, \surf)|\nabla\phi|.
    \label{eq:3d_ls}
}
This replacement enables higher-order shading and visibility gradients while being practical in terms of availability of multiple mesh-based differentiable renderers that are 1)
fast~\cite{laine20,ravi20}, 2) have user-friendly
APIs~\cite{mitsuba3,laine20,li18,ravi20}, and 3) can handle complex light
transport effects~\cite{mitsuba3,li18}.

\paragraph{Large Steps in Visibility Optimization}
Unlike the visibility gradients in present differentiable renderers,
the topological derivative from Result~\ref{res:res3} is defined for the interior of the
visible shape.
For shape recovery from multi-view images, there is the immediate benefit that,
for each point on the surface, we can obtain visibility sensitivities defined
from several viewing angles.
Conversely, with standard shape derivatives, visibility gradients are defined for
points only when they appear at the silhouette.
The resulting optimization signal is quite sparse in practice --- and in cases
with complex lighting and materials, it is noisy.
We find that topological derivatives provide a more robust signal for visibility
changes and enable faster optimization.
In Figure~\ref{fig:large_steps} we show a comparison with visibility gradients
from nvdiffrast~\cite{laine20} for inverse shape recovery from multi-view
images.

\paragraph{Multi-View Reconstruction}
We validate the proposed theory using synthetic 3D shapes of complex topology.
Given a set of images with known material and environment lighting, we use the level-set
evolution in (\ref{eq:3d_ls}) to minimize reprojection error.
We show qualitative and quantitative results in
Figure~\ref{fig:inverse_main} with nvdiffrast~\cite{laine20}.
Since~\cite{laine20} is defined for triangle meshes, we regularize the gradients for smoother optimization using LSIG~\cite{nicolet21} and enable topological changes with NIE~\cite{mehta2022level}.
Theoretically, our method becomes an extension of NIE in the form of an additional
term in the level-set PDE~(\ref{eq:3d_ls}) for topological derivatives.
Starting from a genus-$0$ sphere, we find that
shape derivatives are insufficient to prompt visibility changes in the
shape's interior.
The \texttt{holeball} example, which requires puncturing three axial holes
through the sphere, is a prime illustration of this problem.
The regions of hole nucleation are within the silhouette of the shape and hence
other methods require shading gradients to prompt the visibility changes ---
which, as discussed earlier in Figure~\ref{fig:vis_shade_gradients}, are
unreliable for topology evolution.
\begin{figure}
\begin{minipage}[c]{\textwidth}
    \scriptsize
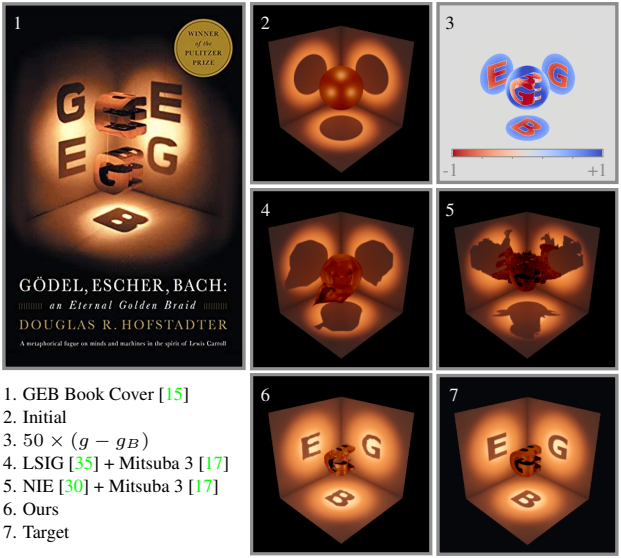
\end{minipage}
    \caption{\textbf{Shape ambigram from a single image.} We reproduce the
    famous shape ambigram from the cover of \emph{GEB} by Hofstader (1). Given a
    single image (7), we aim to recover a shape with complex shadow projections
    from a relatively simple shape (2).
    A differentiable path tracer like Mitsuba 3~\cite{mitsuba3} uses a form of
    shape derivative and struggles with visibility changes necessary in the
    interior regions. We regularize the gradients from~\cite{mitsuba3} using two
    recent methods. One which works with triangle meshes~\cite{nicolet21} (4) and
    the other with level-sets~\cite{mehta2022level} (5). Our method uses 
    topological derivatives for secondary visibility and can successfully
    recover the shadows (6). In (3) we visualize the term $g - g_B$ at
    intialization.
    }
\label{fig:geb_main}
\end{figure}

We find that by using topological derivatives we can successfully recover the
shown examples as they can evaluate visibility changes in such regions.
We provide more examples of shape recovery in the Supplementary.

\subsubsection{Secondary Visibility}
\label{sec:3d_second_vis}
We extend our formulation for TD to a rendering integral with secondary
light bounces. We project a point $\pixel$ on the image plane to a shading point in
the scene by $\x = \pi^{-1}(\pixel)$ (see the inline figure). We can integrate a scene function over the
\setlength\intextsep{0pt}
\setlength{\columnsep}{5pt}%
\begin{wrapfigure}{r}{0pt}
    \small
\begingroup%
  \makeatletter%
  \providecommand\color[2][]{%
    \errmessage{(Inkscape) Color is used for the text in Inkscape, but the package 'color.sty' is not loaded}%
    \renewcommand\color[2][]{}%
  }%
  \providecommand\transparent[1]{%
    \errmessage{(Inkscape) Transparency is used (non-zero) for the text in Inkscape, but the package 'transparent.sty' is not loaded}%
    \renewcommand\transparent[1]{}%
  }%
  \providecommand\rotatebox[2]{#2}%
  \newcommand*\fsize{\dimexpr\f@size pt\relax}%
  \newcommand*\lineheight[1]{\fontsize{\fsize}{#1\fsize}\selectfont}%
  \ifx\svgwidth\undefined%
    \setlength{\unitlength}{47.61527395bp}%
    \ifx\svgscale\undefined%
      \relax%
    \else%
      \setlength{\unitlength}{\unitlength * \real{\svgscale}}%
    \fi%
  \else%
    \setlength{\unitlength}{\svgwidth}%
  \fi%
  \global\let\svgwidth\undefined%
  \global\let\svgscale\undefined%
  \makeatother%
  \begin{picture}(1,1.26489453)%
    \lineheight{1}%
    \setlength\tabcolsep{0pt}%
    \put(0,0){\includegraphics[width=\unitlength,page=1]{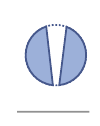}}%
    \put(0.16612541,0.8829188){\makebox(0,0)[lt]{\lineheight{1.25}\smash{\begin{tabular}[t]{l}$\surf$\end{tabular}}}}%
    \put(0,0){\includegraphics[width=\unitlength,page=2]{fig18.pdf}}%
    \put(0.68368229,1.10381351){\makebox(0,0)[lt]{\lineheight{1.25}\smash{\begin{tabular}[t]{l}$g_B$\end{tabular}}}}%
    \put(0.05060048,0.26928994){\makebox(0,0)[lt]{\lineheight{1.25}\smash{\begin{tabular}[t]{l}$\pixel$\end{tabular}}}}%
    \put(0.60809296,0.26544439){\makebox(0,0)[lt]{\lineheight{1.25}\smash{\begin{tabular}[t]{l}$\x'$\end{tabular}}}}%
    \put(0.5275558,0.01852002){\makebox(0,0)[lt]{\lineheight{1.25}\smash{\begin{tabular}[t]{l}$\x$\end{tabular}}}}%
  \end{picture}%
\endgroup%

\end{wrapfigure}

hemispherical domain around this shading point $\x$ and relay the differentials back to
$\pixel$. We follow a similar construction as in the case of primary
visibility (Result~\ref{res:res3}) and estimate TD at a point $\x'\in\surf$ acting as an occluder for the light path $\pixel\to\x\to\x'$. Let $g(\x')$ be the
scene function along this path with the unperturbed surface and $g_B(\x')$ the scene function
after the conical perturbation. The TD of the shading functional around
$\x$ with respect to a perturbation at $\x'$ is,
\eqn{
    D_\tau(\x',\surf) = (g(\x') -
    g_B(\x'))\frac{\kappa(\x'-\x)^t(\x'-\x)}{(\x'-\x)^3_z}.
}
We provide the full derivation in the Supplementary.
A surface evolution equation including a TD of this form can perforate through
shapes along the secondary segments of light paths.
To demonstrate the feasibility of this approach we propose a task of recovering
3D shape ambigrams from a single image.
Inspired by the popular cover of \emph{G\"{o}del, Escher,
Bach}~\cite{hofstadter1982godel}, we construct a scene with an ambigram of
letters
\texttt{G,E} and \texttt{B} placed inside a box with shadow projections of the
letters on different planes.
As shown in Figure~\ref{fig:geb_main}, differentiable path tracers such as Mitsuba
3~\cite{mitsuba3} cannot recover this shape with a genus-$0$ intialization.
By adding a TD term to (\ref{eq:3d_ls}), we can deform the sphere such that
more light can be accumulated in the regions with shadows.

\section{Discussion}
As shown in the experiments (Figures~\ref{fig:sds}, \ref{fig:inverse_main} and
\ref{fig:geb_main}), topological derivatives (TDs) can be used to prompt
visibility changes away from shape contours.
The form of these derivatives resemble occupancy and segmentation loss
functions~\cite{niemeyer2020dvr,yariv20} that are frequently employed in shape
reconstruction methods.
Roughly put, topological derivative can be interpreted as a \emph{soft} notion
of occupancy loss and can be used when a segmentation mask of the final
geometry is unavailable.

In comparison to volumetric-rendering based
methods~\cite{mildenhallNeRFRepresentingScenes2020} the advantage of using TDs
is efficient sampling and the ability to use more complex materials and light
transport effects.
As evident from the synthetic nature of the experiments, however, there is still
a gap between the reconstruction quality achieved using TDs and volumetric
methods.
We anticipate a few significant challenges, that if resolved could reduce the
gap.
\emph{First}, balancing the shape and topological derivative terms in
Eq.~\ref{eq:3d_ls} for shape evolution in 3D is not straightforward. TDs are not
always necessary, especially when the optimization has reached a point where the
topology of the shape is the same as that of the target.
\emph{Second}, the assumption of constant curvature (in
Eq.~\ref{eq:td_3d_final}) could lead to unexpected behavior and its effect needs
a thorough inquiry.
\emph{Third}, we do not derive TDs for phase nucleation in 3D. In 2D, a sample
in the pixel space corresponds to a single sample on the canvas.  As a result,
the term ($g - g_B$) in Equation~\ref{eq:2d_td} relates to a single point in the
shape's exterior. For 3D, in empty space, a pixel-sample corresponds to a ray in
the direction of that sample. This makes the problem of phase nucleation
significantly more challenging.

Finally, joint optimization of geometry and other parameters such color,
material properties and lighting is not trivial and poses additional challenges.
As a preliminary step, below, we show an experiment with the task of recovering
both geometry and color for 2D vectorization.
\vspace{0.5em}
\begin{figure}[h!]
\begin{minipage}[c]{\linewidth}
    \centering
    \footnotesize
\begingroup%
  \makeatletter%
  \providecommand\color[2][]{%
    \errmessage{(Inkscape) Color is used for the text in Inkscape, but the package 'color.sty' is not loaded}%
    \renewcommand\color[2][]{}%
  }%
  \providecommand\transparent[1]{%
    \errmessage{(Inkscape) Transparency is used (non-zero) for the text in Inkscape, but the package 'transparent.sty' is not loaded}%
    \renewcommand\transparent[1]{}%
  }%
  \providecommand\rotatebox[2]{#2}%
  \newcommand*\fsize{\dimexpr\f@size pt\relax}%
  \newcommand*\lineheight[1]{\fontsize{\fsize}{#1\fsize}\selectfont}%
  \ifx\svgwidth\undefined%
    \setlength{\unitlength}{230.39999654bp}%
    \ifx\svgscale\undefined%
      \relax%
    \else%
      \setlength{\unitlength}{\unitlength * \real{\svgscale}}%
    \fi%
  \else%
    \setlength{\unitlength}{\svgwidth}%
  \fi%
  \global\let\svgwidth\undefined%
  \global\let\svgscale\undefined%
  \makeatother%
  \begin{picture}(1,0.22085384)%
    \lineheight{1}%
    \setlength\tabcolsep{0pt}%
    \put(0.06080939,0.19961548){\color[rgb]{0,0,0}\makebox(0,0)[lt]{\lineheight{1.25}\smash{\begin{tabular}[t]{l}Init\end{tabular}}}}%
    \put(0.72794546,0.19961548){\color[rgb]{0,0,0}\makebox(0,0)[lt]{\lineheight{1.25}\smash{\begin{tabular}[t]{l}Est\end{tabular}}}}%
    \put(0.87480727,0.19961548){\color[rgb]{0,0,0}\makebox(0,0)[lt]{\lineheight{1.25}\smash{\begin{tabular}[t]{l}Target\end{tabular}}}}%
    \put(0,0){\includegraphics[width=\unitlength,page=1]{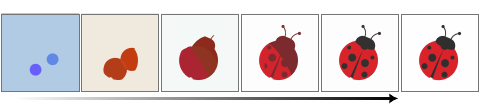}}%
  \end{picture}%
\endgroup%

\end{minipage}
\end{figure}
\vspace{-1em}

\paragraph{Acknowledgements}
This work was supported in part by NSF CAREER 1751365, NSF grant IIS 2110409, a
Qualcomm Innovation Fellowship, gifts from Meta, Adobe and Google, the Ronald
L.  Graham Chair and the UC San Diego Center for Visual Computing. We thank
Martha Gahl and the anonymous reviewers for their suggestions and comments.

\clearpage
\balance
{\small
\bibliographystyle{ieee_fullname}
\bibliography{egbib,zotero}

\begin{thebibliography}{10}\itemsep=-1pt

\bibitem{amstutz2010topological}
Samuel Amstutz and Antonio~A Novotny.
\newblock Topological optimization of structures subject to von mises stress
  constraints.
\newblock {\em Structural and Multidisciplinary Optimization}, 41(3):407, 2010.

\bibitem{bangaru2022differentiable}
Sai Bangaru, Michael Gharbi, Tzu-Mao Li, Fujun Luan, Kalyan Sunkavalli, Milos
  Hasan, Sai Bi, Zexiang Xu, Gilbert Bernstein, and Fredo Durand.
\newblock Differentiable rendering of neural sdfs through reparameterization.
\newblock In {\em ACM SIGGRAPH Asia 2022 Conference Proceedings}, 2022.

\bibitem{bangaru20}
Sai Bangaru, Tzu-Mao Li, and Fr{\'e}do Durand.
\newblock Unbiased warped-area sampling for differentiable rendering.
\newblock {\em ACM Trans. Graph.}, 39(6):245:1--245:18, 2020.

\bibitem{botsch2004remeshing}
Mario Botsch and Leif Kobbelt.
\newblock A remeshing approach to multiresolution modeling.
\newblock In {\em Proceedings of Eurographics/ACM SIGGRAPH symposium on
  Geometry processing}, pages 185--192, 2004.

\bibitem{burgerIncorporatingTopologicalDerivatives2004}
Martin Burger, Benjamin Hackl, and Wolfgang Ring.
\newblock Incorporating topological derivatives into level set methods.
\newblock In {\em Journal of Computational Physics}, 2004.

\bibitem{carpioTopologicalDerivativesShape2008}
Ana Carpio and Maria~Luisa Rapún.
\newblock {\em Topological {{Derivatives}} for {{Shape Reconstruction}}},
  volume 1943 of {\em Lecture {{Notes}} in {{Mathematics}}}.
\newblock 2008.

\bibitem{caselles1995geodesic}
Vicent Caselles, Ron Kimmel, and Guillermo Sapiro.
\newblock Geodesic active contours.
\newblock In {\em Proceedings of IEEE International Conference on Computer
  Vision (ICCV)}, pages 694--699. IEEE, 1995.

\bibitem{chan2001active}
Tony~F Chan and Luminita~A Vese.
\newblock Active contours without edges.
\newblock {\em IEEE Transactions on image processing}, 10(2):266--277, 2001.

\bibitem{chen19}
Zhiqin Chen and Hao Zhang.
\newblock Learning implicit fields for generative shape modeling.
\newblock In {\em Proceedings of the IEEE/CVF Conference on Computer Vision and
  Pattern Recognition (CVPR)}, June 2019.

\bibitem{delaunoyGradientFlowsOptimizing2011}
Amaël Delaunoy and Emmanuel Prados.
\newblock Gradient {{Flows}} for {{Optimizing Triangular Mesh-based Surfaces}}:
  {{Applications}} to {{3D Reconstruction Problems Dealing}} with
  {{Visibility}}.
\newblock In {\em International Journal of Computer Vision}, 2011.

\bibitem{faugerasVariationalPrinciplesSurface}
Olivier Faugeras and Renaud Keriven.
\newblock {\em Variational Principles, Surface Evolution, PDE's, Level Set
  Methods and the Stereo Problem}.
\newblock IEEE, 2002.

\bibitem{gargalloMinimizingReprojectionError2007}
Pau Gargallo, Emmanuel Prados, and Peter Sturm.
\newblock Minimizing the {{Reprojection Error}} in {{Surface Reconstruction}}
  from {{Images}}.
\newblock In {\em Proceedings of IEEE {{International Conference}} on
  {{Computer Vision}} (ICCV)}, 2007.

\bibitem{hales2007jordan}
Thomas~C Hales.
\newblock Jordan’s proof of the jordan curve theorem.
\newblock {\em Studies in logic, grammar and rhetoric}, 10(23):45--60, 2007.

\bibitem{hintermuller2008electrical}
M Hintermuller and Antoine Laurain.
\newblock Electrical impedance tomography: from topology to shape.
\newblock {\em Control and Cybernetics}, 37(4):913--933, 2008.

\bibitem{hofstadter1982godel}
Douglas Hofstader.
\newblock {\em G{\"o}del, Escher, Bach: An Eternal Golden Braid}.
\newblock 1979.

\bibitem{hormander2015analysis}
Lars H{\"o}rmander.
\newblock {\em The analysis of linear partial differential operators I:
  Distribution theory and Fourier analysis}.
\newblock Springer, 2015.

\bibitem{mitsuba3}
Wenzel Jakob, Sébastien Speierer, Nicolas Roussel, Merlin Nimier-David, Delio
  Vicini, Tizian Zeltner, Baptiste Nicolet, Miguel Crespo, Vincent Leroy, and
  Ziyi Zhang.
\newblock Mitsuba 3 renderer, 2022.
\newblock https://mitsuba-renderer.org.

\bibitem{jiang20}
Yue Jiang, Dantong Ji, Zhizhong Han, and Matthias Zwicker.
\newblock Sdfdiff: Differentiable rendering of signed distance fields for 3d
  shape optimization.
\newblock In {\em The IEEE/CVF Conference on Computer Vision and Pattern
  Recognition (CVPR)}, June 2020.

\bibitem{jin2003variational}
Hailin Jin.
\newblock {\em Variational Methods for Shape Reconstruction in Computer
  Vision}.
\newblock Washington University in St. Louis, 2003.

\bibitem{kajiyaRENDERINGEQUATION1986}
James~T Kajiya.
\newblock The rendering equation.
\newblock In {\em Proceedings of the 13th annual conference on Computer
  Graphics and Interactive Techniques}, pages 143--150, 1986.

\bibitem{kass1988snakes}
Michael Kass, Andrew Witkin, and Demetri Terzopoulos.
\newblock Snakes: Active contour models.
\newblock {\em International journal of computer vision}, 1(4):321--331, 1988.

\bibitem{laine20}
Samuli Laine, Janne Hellsten, Tero Karras, Yeongho Seol, Jaakko Lehtinen, and
  Timo Aila.
\newblock Modular primitives for high-performance differentiable rendering.
\newblock {\em ACM Transactions on Graphics}, 39(6), 2020.

\bibitem{larrabide2008topological}
Ignacio Larrabide, RA Feij{\'o}o, AA Novotny, and EA Taroco.
\newblock Topological derivative: a tool for image processing.
\newblock {\em Computers \& Structures}, 86(13-14):1386--1403, 2008.

\bibitem{li18}
Tzu-Mao Li, Miika Aittala, Fr{\'e}do Durand, and Jaakko Lehtinen.
\newblock Differentiable monte carlo ray tracing through edge sampling.
\newblock {\em ACM Trans. Graph. (Proc. SIGGRAPH Asia)}, 37(6):222:1--222:11,
  2018.

\bibitem{liDifferentiableVectorGraphics2020}
Tzu-Mao Li, Michal Lukáč, Michaël Gharbi, and Jonathan Ragan-Kelley.
\newblock Differentiable vector graphics rasterization for editing and
  learning.
\newblock 39(6):1--15.

\bibitem{liu19}
Shichen Liu, Tianye Li, Weikai Chen, and Hao Li.
\newblock Soft rasterizer: A differentiable renderer for image-based 3d
  reasoning.
\newblock {\em The IEEE International Conference on Computer Vision (ICCV)},
  Oct 2019.

\bibitem{liu20}
Shaohui Liu, Yinda Zhang, Songyou Peng, Boxin Shi, Marc Pollefeys, and Zhaopeng
  Cui.
\newblock Dist: Rendering deep implicit signed distance function with
  differentiable sphere tracing.
\newblock In {\em IEEE/CVF Conference on Computer Vision and Pattern
  Recognition (CVPR)}, June 2020.

\bibitem{loper2014opendr}
Matthew~M Loper and Michael~J Black.
\newblock Opendr: An approximate differentiable renderer.
\newblock In {\em Computer Vision--ECCV 2014: 13th European Conference, Zurich,
  Switzerland, September 6-12, 2014, Proceedings, Part VII 13}, pages 154--169.
  Springer, 2014.

\bibitem{loubet2019reparameterizing}
Guillaume Loubet, Nicolas Holzschuch, and Wenzel Jakob.
\newblock Reparameterizing discontinuous integrands for differentiable
  rendering.
\newblock {\em ACM Transactions on Graphics (TOG)}, 38(6):1--14, 2019.

\bibitem{mehta2022level}
Ishit Mehta, Manmohan Chandraker, and Ravi Ramamoorthi.
\newblock A level set theory for neural implicit evolution under explicit
  flows.
\newblock In {\em European Conference on Computer Vision (ECCV)}, pages
  711--729. Springer, 2022.

\bibitem{mildenhallNeRFRepresentingScenes2020}
Ben Mildenhall, Pratul~P Srinivasan, Matthew Tancik, Jonathan~T Barron, Ravi
  Ramamoorthi, and Ren Ng.
\newblock Nerf: Representing scenes as neural radiance fields for view
  synthesis.
\newblock {\em Communications of the ACM}, 65(1):99--106, 2021.

\bibitem{milnor1997topology}
John Milnor and David~W Weaver.
\newblock {\em Topology from the differentiable viewpoint}, volume~21.
\newblock Princeton university press, 1997.

\bibitem{nazarov2003asymptotic}
Serguei~A Nazarov and Jan Soko{\l}owski.
\newblock Asymptotic analysis of shape functionals.
\newblock {\em Journal de Math{\'e}matiques pures et appliqu{\'e}es},
  82(2):125--196, 2003.

\bibitem{newcombe2011kinectfusion}
Richard~A Newcombe, Shahram Izadi, Otmar Hilliges, David Molyneaux, David Kim,
  Andrew~J Davison, Pushmeet Kohi, Jamie Shotton, Steve Hodges, and Andrew
  Fitzgibbon.
\newblock Kinectfusion: Real-time dense surface mapping and tracking.
\newblock In {\em 2011 10th IEEE international symposium on mixed and augmented
  reality}, pages 127--136. Ieee, 2011.

\bibitem{nicolet21}
Baptiste Nicolet, Alec Jacobson, and Wenzel Jakob.
\newblock Large steps in inverse rendering of geometry.
\newblock {\em ACM Transactions on Graphics (Proceedings of SIGGRAPH Asia)},
  40(6), Dec. 2021.

\bibitem{niemeyer2020dvr}
Michael Niemeyer, Lars Mescheder, Michael Oechsle, and Andreas Geiger.
\newblock Differentiable volumetric rendering: Learning implicit 3d
  representations without 3d supervision.
\newblock In {\em Proceedings of the IEEE/CVF Conference on Computer Vision and
  Pattern Recognition (CVPR)}, pages 3504--3515, 2020.

\bibitem{novotny2019applications}
Antonio~Andr{\'e} Novotny, Jan Soko{\l}owski, and Antoni {\.Z}ochowski.
\newblock {\em Applications of the topological derivative method}.
\newblock Springer, 2019.

\bibitem{osher88}
Stanley Osher and James~A Sethian.
\newblock Fronts propagating with curvature-dependent speed: Algorithms based
  on hamilton-jacobi formulations.
\newblock {\em Journal of computational physics}, 1988.

\bibitem{park19}
Jeong~Joon Park, Peter Florence, Julian Straub, Richard Newcombe, and Steven
  Lovegrove.
\newblock Deepsdf: Learning continuous signed distance functions for shape
  representation.
\newblock In {\em Proceedings of the IEEE/CVF Conference on Computer Vision and
  Pattern Recognition (CVPR)}, June 2019.

\bibitem{poole2022dreamfusion}
Ben Poole, Ajay Jain, Jonathan~T Barron, and Ben Mildenhall.
\newblock Dreamfusion: Text-to-3d using 2d diffusion.
\newblock {\em arXiv preprint arXiv:2209.14988}, 2022.

\bibitem{ravi20}
Nikhila Ravi, Jeremy Reizenstein, David Novotny, Taylor Gordon, Wan-Yen Lo,
  Justin Johnson, and Georgia Gkioxari.
\newblock Accelerating 3d deep learning with pytorch3d.
\newblock {\em arXiv:2007.08501}, 2020.

\bibitem{remelli20}
Edoardo Remelli, Artem Lukoianov, Stephan Richter, Benoit Guillard, Timur
  Bagautdinov, Pierre Baque, and Pascal Fua.
\newblock Meshsdf: Differentiable iso-surface extraction.
\newblock In {\em Advances in Neural Information Processing Systems}, pages
  22468--22478. Curran Associates, Inc., 2020.

\bibitem{rombach2022high}
Robin Rombach, Andreas Blattmann, Dominik Lorenz, Patrick Esser, and Bj{\"o}rn
  Ommer.
\newblock High-resolution image synthesis with latent diffusion models.
\newblock In {\em Proceedings of the IEEE/CVF Conference on Computer Vision and
  Pattern Recognition (CVPR)}, pages 10684--10695, 2022.

\bibitem{sokolowski1999topological}
Jan Sokolowski and Antoni Zochowski.
\newblock On the topological derivative in shape optimization.
\newblock {\em SIAM journal on control and optimization}, 37(4):1251--1272,
  1999.

\bibitem{solemGeometricFormulationGradient2005}
Jan~Erik Solem and Niels~Chr. Overgaard.
\newblock A {{Geometric Formulation}} of {{Gradient Descent}} for {{Variational
  Problems}} with {{Moving Surfaces}}.
\newblock In {\em Scale {{Space}} and {{PDE Methods}} in {{Computer Vision}}},
  volume 3459 of {\em Lecture {{Notes}} in {{Computer Science}}}, pages
  419--430. 2005.

\bibitem{vicini2022differentiable}
Delio Vicini, S{\'e}bastien Speierer, and Wenzel Jakob.
\newblock Differentiable signed distance function rendering.
\newblock {\em ACM Transactions on Graphics (TOG)}, 41(4):1--18, 2022.

\bibitem{wang2021neus}
Peng Wang, Lingjie Liu, Yuan Liu, Christian Theobalt, Taku Komura, and Wenping
  Wang.
\newblock Neus: Learning neural implicit surfaces by volume rendering for
  multi-view reconstruction.
\newblock {\em arXiv preprint arXiv:2106.10689}, 2021.

\bibitem{whitaker1998level}
Ross~T Whitaker.
\newblock A level-set approach to 3d reconstruction from range data.
\newblock {\em International journal of computer vision}, 29(3):203--231, 1998.

\bibitem{yariv2021volume}
Lior Yariv, Jiatao Gu, Yoni Kasten, and Yaron Lipman.
\newblock Volume rendering of neural implicit surfaces.
\newblock In {\em Advances in Neural Information Processing Systems},
  volume~34, 2021.

\bibitem{yariv20}
Lior Yariv, Yoni Kasten, Dror Moran, Meirav Galun, Matan Atzmon, Basri Ronen,
  and Yaron Lipman.
\newblock Multiview neural surface reconstruction by disentangling geometry and
  appearance.
\newblock In {\em Advances in Neural Information Processing Systems},
  volume~33, 2020.

\bibitem{zhang20}
Cheng Zhang, Bailey Miller, Kai Yan, Ioannis Gkioulekas, and Shuang Zhao.
\newblock Path-space differentiable rendering.
\newblock {\em ACM Trans. Graph.}, 39(4):143:1--143:19, 2020.

\bibitem{zhao1996variational}
Hong-Kai Zhao, Tony Chan, Barry Merriman, and Stanley Osher.
\newblock A variational level set approach to multiphase motion.
\newblock {\em Journal of computational physics}, 127(1):179--195, 1996.

\end{thebibliography}
}
\end{document}